\pdfoutput=1

\documentclass[11pt]{article}
\usepackage{amsmath}
\usepackage[rgb]{xcolor}
\usepackage{amssymb} 
\usepackage{arydshln} 

\usepackage{acl}
\usepackage{graphicx}
\usepackage{siunitx}
\usepackage{lipsum}
\usepackage{times}
\usepackage{latexsym}
\usepackage{booktabs}
\usepackage{adjustbox}
\usepackage{multirow}
\usepackage[T1]{fontenc}
\usepackage{enumitem}
\usepackage{listings}
\usepackage[utf8]{inputenc}
\usepackage[linesnumbered,ruled]{algorithm2e}
\usepackage{microtype}
\usepackage{tabularx}
\usepackage{authblk}

\usepackage{tikz}
\usepackage{inconsolata}
\usepackage{tabularx}

\usepackage{amsmath}
\definecolor{customPink}{RGB}{248,234,233}
\definecolor{customYellow}{RGB}{247,235,206}
\definecolor{customGreen}{RGB}{212,255,191}
\definecolor{customRed}{RGB}{255,194,212}
\title{Joint Visual and Text Prompting for Zero-Shot Object-Oriented Perception with Multimodal Large Language Models}

\author[1]{Songtao Jiang\thanks{Equal contribution}}
\author[2]{Yan Zhang\textsuperscript{*}}
\author[1]{Chenyi Zhou}
\author[2]{Yeying Jin}
\author[3]{Yang Feng}
\author[1]{Jian Wu}
\author[1]{Zuozhu Liu}

\affil[1]{Zhejiang University}
\affil[2]{National University of Singapore}
\affil[3]{Angelalign Inc., China}

\begin{document}
\maketitle
\begin{abstract}

Multimodal Large Language Models (MLLMs) such as GPT-4V and Gemini Pro face challenges in achieving human-level perception in Visual Question Answering (VQA), particularly in object-oriented perception tasks which demand fine-grained understanding of object identities, locations or attributes, as indicated by empirical findings. This is mainly due to their limited capability to effectively integrate complex visual cues with textual information and potential object hallucinations. In this paper, we present a novel approach, Joint Visual and Text Prompting (VTPrompt), that employs fine-grained visual information to enhance the capability of MLLMs in VQA, especially for object-oriented perception. VTPrompt merges visual and text prompts to extract key concepts from textual questions and employs a detection model to highlight relevant objects as visual prompts in images. The processed images alongside text prompts are subsequently fed into MLLMs to produce more accurate answers. Our experiments with  GPT-4V and Gemini Pro, on three benchmarks, i.e., MME , MMB and POPE, demonstrate significant improvements. Particularly, our method led to a score improvement of up to 183.5 for GPT-4V on MME and enhanced MMB performance by 8.17\% for GPT-4V and 15.69\% for Gemini Pro. Our code is released on \href{https://github.com/jiangsongtao/VTprompt}{\texttt{https://github.com/jiangsongtao/VTprompt}}.


\end{abstract}

\section{Introduction}

\begin{figure}[t!]
    \centering
    \includegraphics[width=1\linewidth]{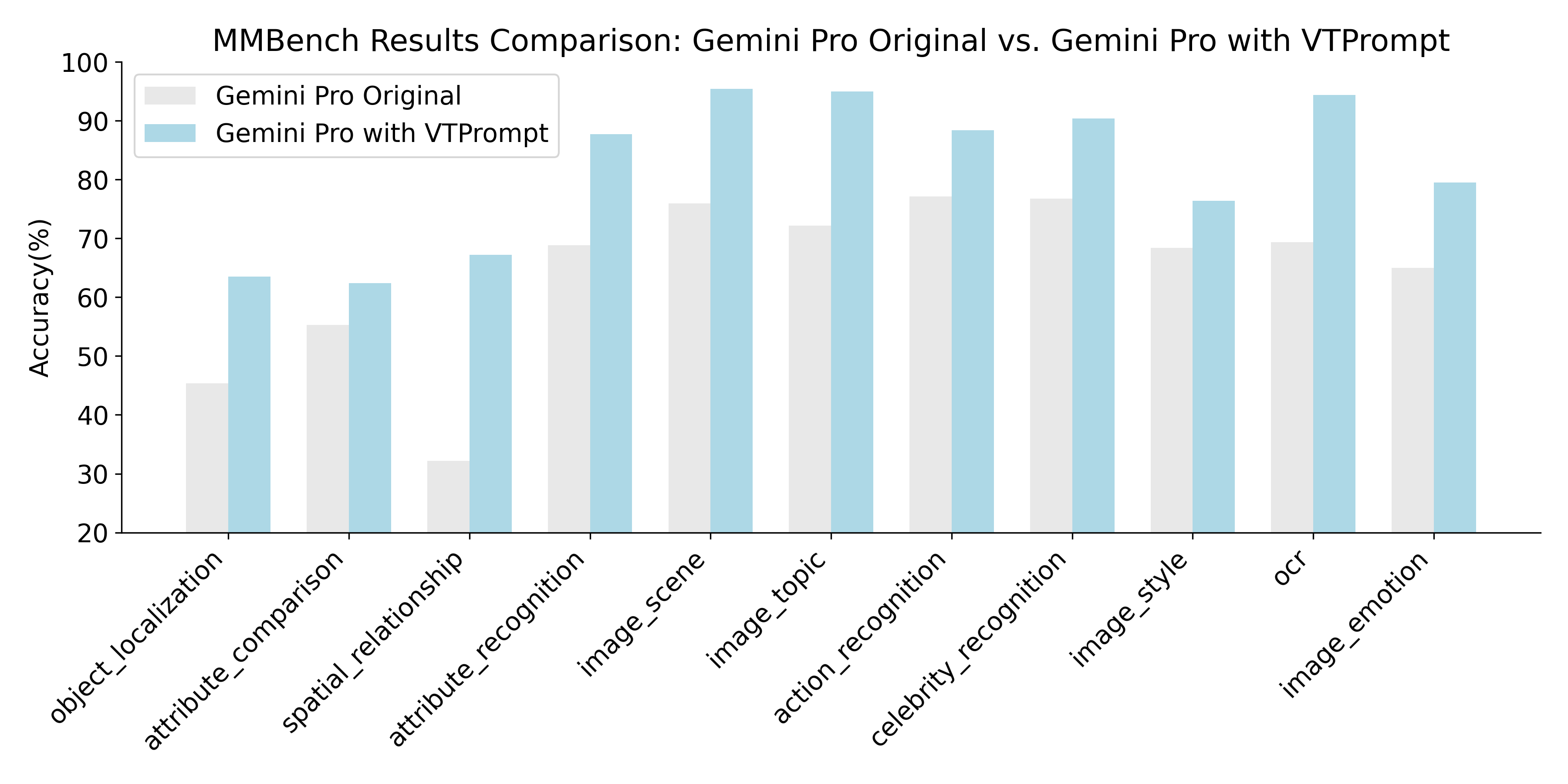}
    \caption{Performance of Gemini Pro ~\citep{geminiteam2023gemini} on MMB ~\citep{liu2023mmbench}. The inferior performance on the three object-oriented tasks (left-most) can be boosted with our VTPrompt.We also present the results based on GPT-4V in the Appendix Figure \ref{fig:mmb4v}.}
    \label{fig:mmb}
\end{figure}

A long-term yet challenging goal in AI is to achieve human-level perception with multimodal vision and textual information~\citep{miao2022research,goyal2016towards,de2023visual,antol2015vqa}. In this endeavor, the Visual Question Answering (VQA) task stands out as a pivotal benchmark, which evaluates the ability of AI systems to analyze and interpret both visual and textual information to generate responses ~\citep{wu2017visual}. Recently, Multimodal Large Language Models (MLLMs), such as  GPT-4V~\citep{OpenAI2023GPT4V}, Gemini Pro~\citep{geminiteam2023gemini}, LLaVA~\citep{liu2023visual} and MiniGPT4-V2~\citep{chen2023minigptv2}, have demonstrated promising capability in VQA tasks. However, our empirical evaluations in Figure \ref{fig:mmb} about Gemini Pro in MMB's multimodal perception tasks indicate their inferior performance on object-oriented tasks, such as object localization, spatial relationships, and attribute comparison. Consistent findings of GPT4V as shown in the Appendix highlight a specific weakness of MLLMs in handling object-oriented tasks.

\begin{figure*}[t!]
    \centering
    \includegraphics[width=1\linewidth]{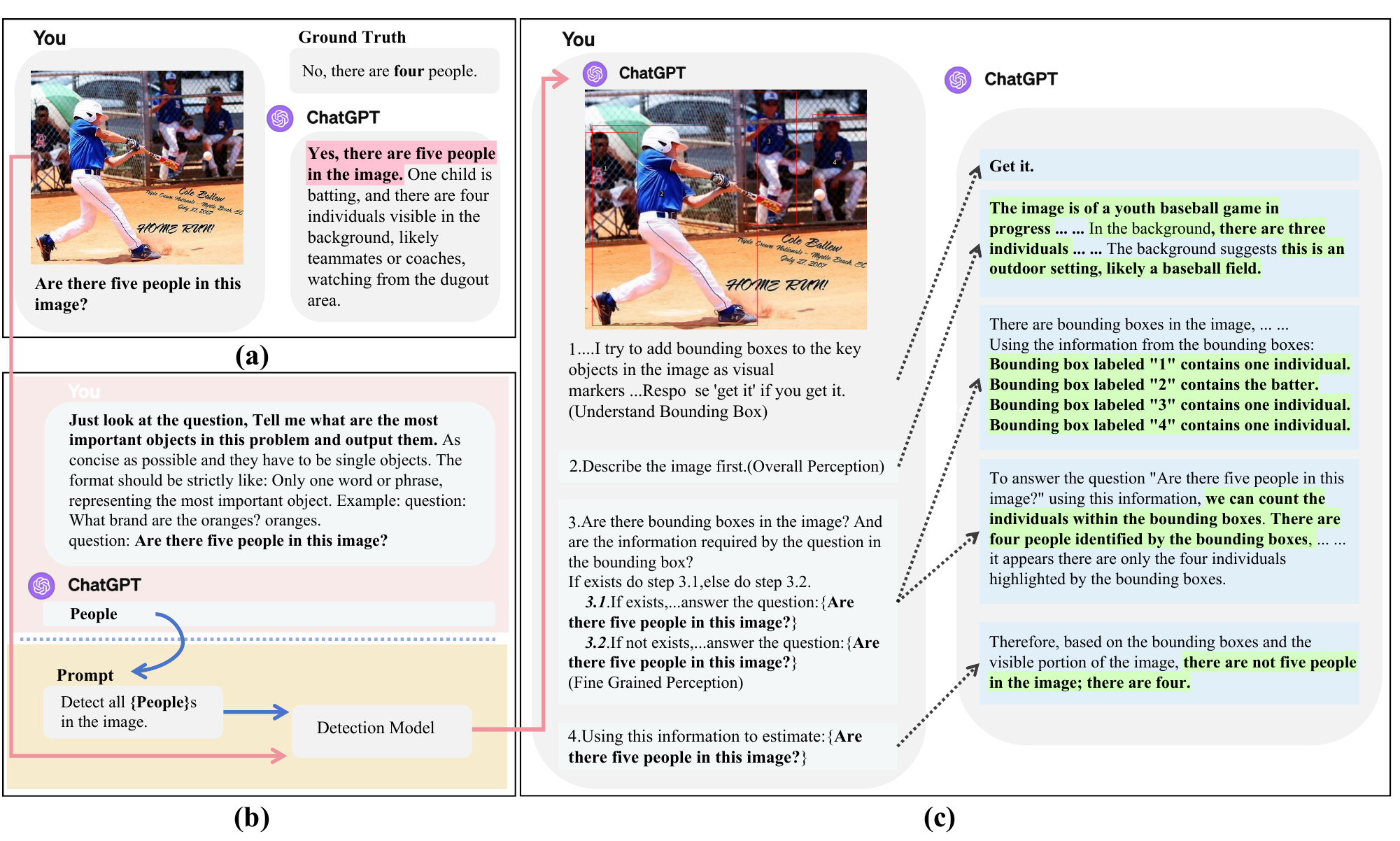}
        \caption{ \textbf{(a)} Regular VQA with GPT-4V generating wrong answers. \textbf{(b-c)} Pipeline of our VTPrompt. The \protect\tikz[baseline=-0.5ex]\protect\draw[fill=customPink] (0,0) rectangle (0.5,0.2); represents the \textbf{Key Concepts Extraction}, corresponding to Section \ref{sec:2.21}, and the \protect\tikz[baseline=-0.5ex]\protect\draw[fill=customYellow] (0,0) rectangle (0.5,0.2); illustrates the \textbf{VPrompt Generation}, as detailed in Section \ref{sec:2.3}. The generated image with visual markers from \textbf{(b)} are processed in \textbf{(c)} which focuses on  \textbf{TPrompt for Answer Generation} as in Section \ref{sec:2.4}, where the image enhanced with visual and text prompts are combined and fed into GPT-4V to produce the answers, as indicated by \protect\tikz[baseline=-0.5ex]\protect\draw[fill=customGreen] (0,0) rectangle (0.5,0.2);.}
    \label{fig:Pipeline}
    \vspace{-5mm}
\end{figure*}

Resolving object-oriented tasks with MLLMs remain challenging. On one hand, existing MLLMs usually experience difficulties for effective and accurate visual grounding and interpretation~\citep{yuan2021perception}, e.g., they may not be able to accurately find or locate the critial objects which are essential to correctly answer the question, as shown in Figure~\label{fig:case1} in the Appendix.  On the other hand, there's a tendency towards object hallucination~\citep{liu2023hallusionbench,chen2024unified,zheng2024exploring,yin2023woodpecker}, where MLLMs might perceive objects that aren't present, compounding the difficulty in achieving precise object-oriented perception. For instance, as shown in Figure~\ref{fig:Pipeline}(a), the MLLM failed to count the number of persons in the image due to wrong object recognition.

In this paper, we introduce VTPrompt, a novel approach that significantly enhances MLLMs' object-oriented perception by  integrating both visual and textual prompts. As illustrated in Figure~\ref{fig:Pipeline}, VTPrompt first extract key concepts from the textual question, which are employed to guide a detection model, e.g., SPHINX~\citep{lin2023sphinx} or SAM~\citep{kirillov2023segment}, for precise object marking. This not only ensures accurate localization but also enriches the model's interpretive capabilities through text prompts that encapsulate the question, refined with visual cues. The refined image with visual markers are fed into the MLLMs, which are further guided by the optimized text prompts to grab meaningful understandings of the bounding boxes, the overall image, as well as fine-grained object perception to obtain the final answer.  We evaluate VTPrompt on three popular benchmarks, i.e., MME\citep{fu2023mme} , MMB\citep{liu2023mmbench}, POPE\citep{li2023evaluating} with top-performing MLLMs, i.e.,  GPT-4V and Gemini Pro,  demonstrated consistent performance enhancements. Notably, there was an increase of up to 183.5 points on the MME benchmark for GPT-4V, which is known for its complexity. Additionally, performance on MMB was enhanced by 8.17\% for GPT-4V and 15.69\% for Gemini Pro, which established new state-of-the-art performance  on MMB. 

\begin{figure}[t!]
    \centering
    \includegraphics[width=1\linewidth]{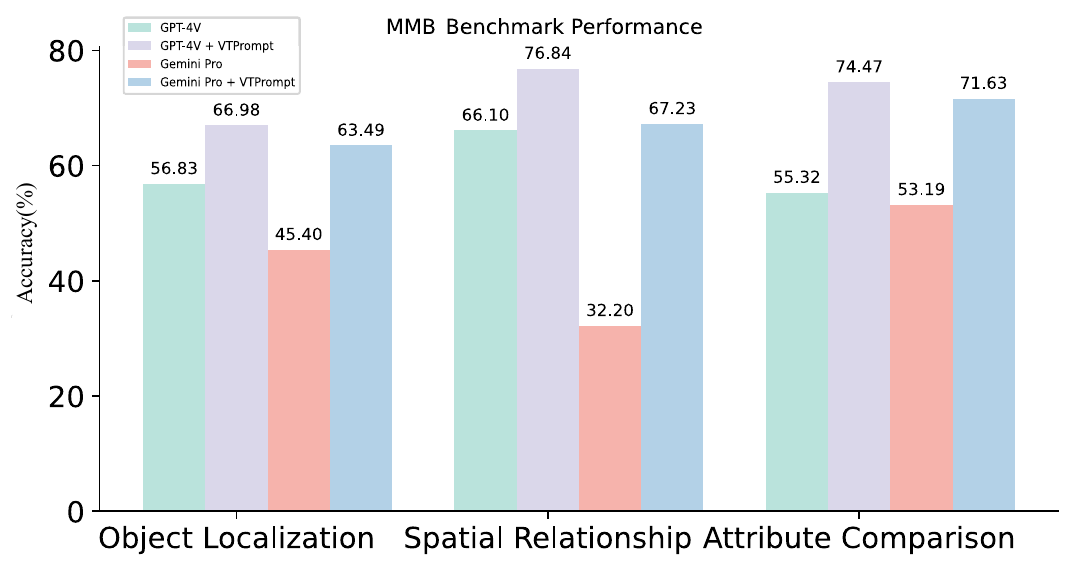}
    \includegraphics[width=1\linewidth]{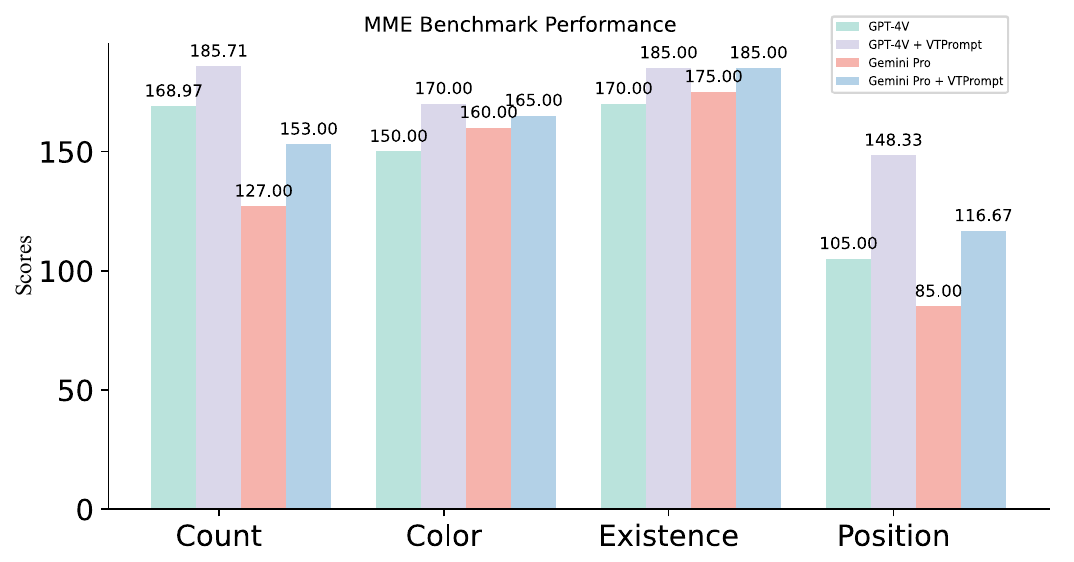}
    \caption{Performance of GPT-4V and Gemini Pro on Object-Oriented Perception Tasks in MMB and MME Benchmarks.}
    \label{fig:tableMMBench_MME}
\end{figure}

\section{Method}
\subsection{VQA}
\label{sec:2.1}
Visual Question Answering (VQA) can be formally defined as a function \(f\) that maps a given image-question pair \((I, Q)\) to an answer \(A\),
\begin{equation}
    A = f(I, Q),
\end{equation}
where \(I\) represents the input image, \(Q\) is the natural language question related to the image, and \(A\) is the answer generated by the model. This task requires the model to not only perceive visual elements within the image but also to understand and process the question's semantic content to generate a relevant and accurate answer.


Our VTprompt mainly includes three components including Key Concepts Extraction,  Visual Prompt Generation and Text Prompt for Answer Generation. For simplicity, visual prompts are referred to as VPrompt, and text prompts as TPrompt.
\subsection{Key Concepts Extraction}
\label{sec:2.21}
VTprompt first feeds the text question from the VQA task into GPT-4\citep{openai2023gpt}, which, following the prompt structure displayed in Figure \ref{fig:Pipeline} (b), extracts the key objects. For straightforward object-related questions, GPT-4 generally can identify the key objects directly. In contrast, for complex questions that demand an understanding of the entire scene, GPT-4 often suggest 3-5 objects that are most relevant to the context. These key concepts extracted from the question signify the elements within the image that require the model's attention for accurate perception and response,
\begin{equation}
    C = \text{Extract}(Q, P_{E}),
\end{equation}
where \( C \) represents the key concepts from question \( Q \), and \( P_{E} \) denotes the extraction-oriented prompt used to guide GPT-4's concept extraction process as detailed in Figure \ref{fig:Pipeline} (b).

\subsection{VPrompt Generation}
\label{sec:2.3}
After obtaining the key objects from the key concept extraction phase, we utilize them as prompts alongside the original image to engage SPHINX\citep{lin2023sphinx}, a model with robust zero-shot detection capabilities, to produce an image annotated with visual markers, specifically bounding boxes that highlight the key objects identified. This visually marked image serves to cue the model to prioritize these regions, enhancing the focus on object information for the VQA task,
\begin{equation}
    I' = \text{SPHINX}(I, C),
\end{equation}
Here, \(I'\) is the image with visual markers, \(I\) is the original VQA image,  and \(C\) denotes the set of key concepts to be detected, as detailed in Figure \ref{fig:Pipeline} (b).

\subsection{TPrompt for Answer Generation}
\label{sec:2.4}
To ensure effective utilization of visual markers for enhancing object-oriented perception in VQA, we develope a structured text prompt as shown in Figure \ref{fig:prompt_template}. The prompt, illustrated in the Figure \ref{fig:Pipeline} (c), comprises three critical steps: informing the model about the presence of visual markers without being distracted by their attributes; establishing an overall scene perception to avoid missing environmental cues; prompting the model to discern detailed fine-grained information within visual markers and confirming the relevance of marked objects to the question. This strategy aims to guide MLLMs in generating accurate visual chains of thought by leveraging marked visual information alongside the original question,
\begin{equation}
    A = \text{Model}(I', P_T(Q)),\
\end{equation}
The formula encapsulates the answer generation process where \(A\) is the answer produced by the model. This model processes the image \(I'\) that has been annotated with visual markers, alongside a text prompt \(P_T\), which is tailored based on the original question \(Q\). 




\begin{table*}[t!]
\centering
\small
\resizebox{\textwidth}{!}{%
\begin{tabular}{llllrr}
\toprule
\textbf{Model} & \textbf{Params} & \textbf{Language Model} & \textbf{Vision Model} & \textbf{MME Total} & \textbf{MMB(Dev)} \\
\midrule
\addlinespace
\multicolumn{6}{l}{\textit{\textbf{Baseline Models}}} \\
\midrule
Qwen-VL-Plus\citep{bai2023qwen} & - & QwenLM & - & 2183.30 & 67.00 \\
LLaVA-v1.5-13B\citep{liu2023improved} & 13.4B & Vicuna-v1.5-13B & CLIP ViT-L/14 & 1531.30 & 67.80 \\
ShareGPT4V-13B\citep{chen2023sharegpt4v} & 13.4B & Vicuna-v1.5-13B & CLIP ViT-L/14 & 1921.90 & 68.90 \\
InternLM-XComposer-VL\cite{zhang2023internlm} & 8B & InternLM-7B & EVA-G & 1919.40 & 74.40 \\
LLaVA-InternLM2-20B (QLoRA)\citep{dong2024internlm} & 20.2B & InternLM2-20B & CLIP ViT-L/14 & 1868.00 & 69.00 \\
InternLM-XComposer2-VL & 7B & InternLM2 & CLIP ViT-L/14 & 2243.50 & 79.60\\
Qwen-VL-Max & - & QwenLM & - & \textbf{2433.50} & 77.60 \\
GPT-4V\citep{OpenAI2023GPT4V} & - & - & - & 1926.50 & 77.00 \\
Gemini-Pro\citep{geminiteam2023gemini} & - & - & - & 1933.30 & 65.00 \\
\midrule
\addlinespace
\multicolumn{6}{l}{\textit{\textbf{With VTprompt}}} \\
\midrule
GPT-4V+VTprompt & - & - & - & 2110.00(\textbf{+183.50}) & \textbf{85.17(+8.17)} \\
Gemini-Pro+VTprompt & - & - & - & 2054.00(\textbf{+120.70}) & 80.69(\textbf{+15.69}) \\
\bottomrule
\end{tabular}
}
\caption{Overall Model Performances on MME and MMB}
\label{tab:model_summary}
\end{table*}

\section{Experiement Setup}

\noindent \textbf{MME} \citep{fu2023mme} is a comprehensive dataset for evaluating multimodal large language models (MLLMs), focusing on both perception and cognition through 14 distinct subtasks. It uniquely features manually designed instruction-answer pairs to prevent data leakage from public dataset usage. The MME framework assesses model performance via accuracy (Acc) and accuracy+ (Acc+), contributing to an overall score of 2800 points across all subtasks. The scoring formulas are as follows:
\begin{equation}
\text{Acc} = \left( \frac{\text{Correctly Answered Questions}}{\text{Total Questions}} \right) \times 100
\end{equation}
\begin{equation}
\small
\text{Acc+} = \left( \frac{\text{Images Correctly Answered}}{\text{Total Images}} \right) \times 100
\end{equation}
The final score combines these metrics, with each subtask offering up to 200 points:
\begin{equation}
\text{Score} = \sum (\text{Acc} + \text{Acc+})
\end{equation}

\noindent \textbf{MMB} \citep{liu2023mmbench} is a systematically-designed objective benchmark for robustly evaluating the various abilities of multimodal large language models (MLLMs), covering 20 different ability dimensions. Consistent with the original evaluation criteria, our metrics focus on calculating accuracy as the ratio of correctly answered questions to the total number of questions, presented through the following formula:
\begin{equation}
\text{Acc} = \left( \frac{\text{Correctly Answered Questions}}{\text{Total Questions}} \right) \times 100
\end{equation}

\noindent \textbf{POPE} The Polling-based Object Probing Evaluation (POPE)~\citep{li2023evaluating} focuses on object-oriented perception tasks, evaluating hallucination in Language-Vision Models through queries on the presence of specific objects in images. It balances queries between existent and non-existent objects (50\% each) across three settings: in the random setting, objects absent from the image are chosen randomly. The popular setting selects missing objects from a high-frequency pool, while in the adversarial setting, co-occurring objects not present in the image are prioritized. Each setting random samples 100 instances due to API constraints, offering a streamlined evaluation of object-oriented perception tasks.

\noindent \textbf{Object-Oriented Tasks} This kind of tasks involves MLLMs analyzing and responding to questions about objects' identities, locations, and attributes in images. These tasks, essential for evaluating MLLMs in VQA, focus on understanding detailed object characteristics and their contextual relationships within visual content. In the MMB dataset, such tasks include object localization (finding and identifying objects), spatial relationships (determining how objects are positioned relative to each other), and attribute comparison (comparing object features). Similarly, in the MME dataset, counting (how many objects are present), existence (whether an object exists), color (identifying colors of objects), and position (where objects are located) are considered object-oriented tasks.

\noindent \textbf{Baselines} We chose GPT-4V and Gemini Pro as our baselines, representing the forefront of MLLMs. In addition, we compare our approach with current SOTA open-sourced MLLMs on MMB and MME datasets, such as Qwen-VL-Max \citep{bai2023qwen} and InternLM-XComposer2-VL \citep{zhang2023internlm}, which have achieved leading performance through advanced pre-training techniques. 

\noindent \textbf{Experimental Settings} For key concept extraction from VQA questions, we use GPT-4's API, setting the temperature to 0 and max tokens to 2048 for stable results. Answers via GPT-4V and Gemini Pro employ their respective APIs (gpt-4-vision-preview and gemini-pro-vision), both with the temperature at 0 and max tokens at 2048 to ensure consistency and prevent answer truncation.

\section{Main Results}
\begin{table*}[t!]
\centering
\small
\begin{tabular}{
  @{} l 
  l 
  *{4}{S[table-format=3.1]} 
  @{}}
\toprule
\textbf{Setting} & \textbf{Method} & {\textbf{Accuracy}} & {\textbf{Precision}} & {\textbf{Recall}} & {\textbf{F1 Score}} \\
\midrule
\multirow{4}{*}{Adversarial} 
& GPT-4V                 & 79.0 & 92.1 & 66.0 & 76.9 \\
& GPT-4V+VTprompt        & \textbf{82.0(+3.0)} & \textbf{97.3(+5.2)} & \textbf{67.9(+1.9)} & \textbf{80.0(+3.1)} \\
\cdashline{2-6}
& Gemini Pro             & 80.0 & 90.2 & 69.8 & 78.7 \\
& Gemini Pro+VTprompt & \textbf{87.0(+7.0)} & \textbf{93.5(+3.3)} & \textbf{81.1(+11.3)} & \textbf{86.9(+8.2)} \\
\midrule
\multirow{4}{*}{Popular} 
& GPT-4V                 & 78.0 & 87.8 & 67.9 & 76.6 \\
& GPT-4V+VTprompt        & \textbf{85.0(+7.0)} & \textbf{97.5(+9.7)} & \textbf{73.6(+5.7)} & \textbf{83.9(+7.3)} \\
\cdashline{2-6}
& Gemini Pro             & 82.0 & 90.7 & 73.6 & 81.2 \\
& Gemini Pro+VTprompt & \textbf{86.0(+4.0)} & \textbf{93.3(+2.6)} & \textbf{79.2(+5.6)} & \textbf{85.7(+4.5)} \\
\midrule
\multirow{4}{*}{Random} 
& GPT-4V                 & 73.0 & 96.9 & 54.4 & 69.7 \\
& GPT-4V+VTprompt        & \textbf{81.0(+8.0)} & \textbf{100.0(+3.1)} & \textbf{66.7(+12.3)} & \textbf{80.0(+10.3)} \\
\cdashline{2-6}
& Gemini Pro             & 85.0 & 95.7 & 77.2 & 85.4 \\
& Gemini Pro+VTprompt & \textbf{88.0(+3.0)} & \textbf{97.9(+2.2)} & \textbf{80.7(+3.5)} & \textbf{88.5(+3.1)} \\
\bottomrule
\end{tabular}
\caption{Sampled POPE performance for GPT-4V and Gemini Pro across different settings.}
\label{tab:POPE_results}
\end{table*}

\begin{table*}[t!]
\centering
\small
\resizebox{0.9\textwidth}{!}{%
\begin{tabular}{lcccccccc}
\toprule
 & \multicolumn{4}{c}{\textbf{GPT-4V with VP}} & \multicolumn{4}{c}{\textbf{Gemini Pro with VP}} \\
\cmidrule(lr){2-5} \cmidrule(lr){6-9}
\textbf{Task} & \textbf{No TP} & \textbf{+ ZS CoT} & \textbf{+ L-to-M} & \textbf{+ TP} & \textbf{No TP} & \textbf{+ ZS CoT} & \textbf{+ L-to-M} & \textbf{+ TP} \\
\midrule
Object Localization(\%) & 57.74 & 60.31 & 61.90 & \textbf{66.98} & 47.62 & 55.56 & 52.38 & \textbf{63.49} \\
Spatial Relationship(\%) & 67.23 & 70.36 & 69.49 & \textbf{76.84} & 42.54 & 60.45 & 53.67 & \textbf{67.23} \\
Attribute Comparison(\%) & 56.74 & 67.38 & 61.70 & \textbf{74.47} & 57.34 & 63.83 & 59.57 & \textbf{71.63} \\
\bottomrule
\end{tabular}
}
\caption{Ablation study on text prompts in MMB Benchmark. "VP" stands for VPrompt; "ZS CoT" represents Zero-shot Chain of Thought; "L-to-M" signifies Least-to-Most reasoning; "TP" denotes TPrompts.}
\label{tab:text_prompts_experiments_transposed}
\end{table*}

\noindent \textbf{Results on Object-Oriented Perception Subset of MMB}
As shown in Figure \ref{fig:tableMMBench_MME}, our experiments demonstrate that VTPrompt significantly boosts the performance of both GPT-4V and Gemini Pro across object localization, spatial relationships, and attribute comparison tasks. With VTPrompt, GPT-4V's performance increased by 10.15\% in object localization, 10.74\% in spatial relationships, and 19.15\% in attribute comparison. Similarly, Gemini Pro shows substantial improvements: 18.09\% in object localization, 35.03\% in spatial relationships, and 16.31\% in attribute comparison. These improvements highlight VTPrompt's ability to refine models' capacity for detailed object analysis.

\noindent \textbf{Results on Object-Oriented Perception Subset of MME}
On the MME benchmark shown in Figure \ref{fig:tableMMBench_MME}, focused on counting, color identification, existence verification, and positional understanding, VTPrompt again proved its efficacy. 
GPT-4V enhance its scores by 16.74 points in counting, 20 points in color, 15 points in existence, and a remarkable 43.33 points in position. 
For Gemini Pro, notable gains were observed with 26 points in counting, 5 points in color, 10 points in existence, and 31.67 points in position, demonstrating VTPrompt's impact on improving object-oriented task handling.

\noindent \textbf{Overall Performance of MMB}
Our evaluation on the full MMB dataset assesses MLLMs across a range of tasks, beyond object-oriented perception, to include multimodal reasoning and cognition perception. The integration of VTPrompt into GPT-4V and Gemini Pro significantly boost their MMB performance, as detailed in Table~\ref{tab:model_summary}.

With VTPrompt, GPT-4V's score improved substantially, achieving state-of-the-art (SOTA) performance on the MMB dataset. This underscores VTPrompt's effectiveness in enhancing multimodal comprehension and reasoning. Similarly, Gemini Pro saw marked performance gains, demonstrating VTPrompt's capacity to elevate MLLMs' complex data processing and reasoning abilities.

\noindent \textbf{Overall Performance of MME}
We evaluate GPT-4V and Gemini Pro on the entire MME dataset to gauge their overall capabilities, focusing on both perception and reasoning abilities. The combination of our visual and text prompting method has notably enhanced their performance across the board, as detailed in Table~\ref{tab:model_summary}.

Qwen-VL-Max achieve the highest baseline MME score of 2433.50. The application of VTPrompt led to a substantial increase in performance: GPT-4V's score improved by 183.50 points to 2110.00, and Gemini Pro's score increased by 120.70 points to 2054.00. These results underscore the dual impact of VTPrompt, which not only enhances object-oriented perception but also amplifies the overall perception and reasoning capabilities of MLLMs.

\noindent \textbf{Results on POPE}
In the POPE, VTprompt significantly improve the performance of GPT-4V and Gemini Pro models in various settings as shown in Figure \ref{tab:POPE_results}. For the "Adversarial" scenario, Gemini Pro with VTprompt saw accuracy increase to 87.0, with precision, recall, and F1 Score rising to 93.5, 81.1, and 86.9, respectively. Conversely, GPT-4V with VTprompt excelled in the "Random" setting, boosting accuracy to 81.0 and precision to 100.0, with recall and F1 Score enhancing to 66.7 and 80.0. These advancements highlight VTprompt's effectiveness in enhancing object-oriented perception and reducing hallucinations.

\begin{figure}[t!]
    \centering
    \includegraphics[width=1\linewidth]{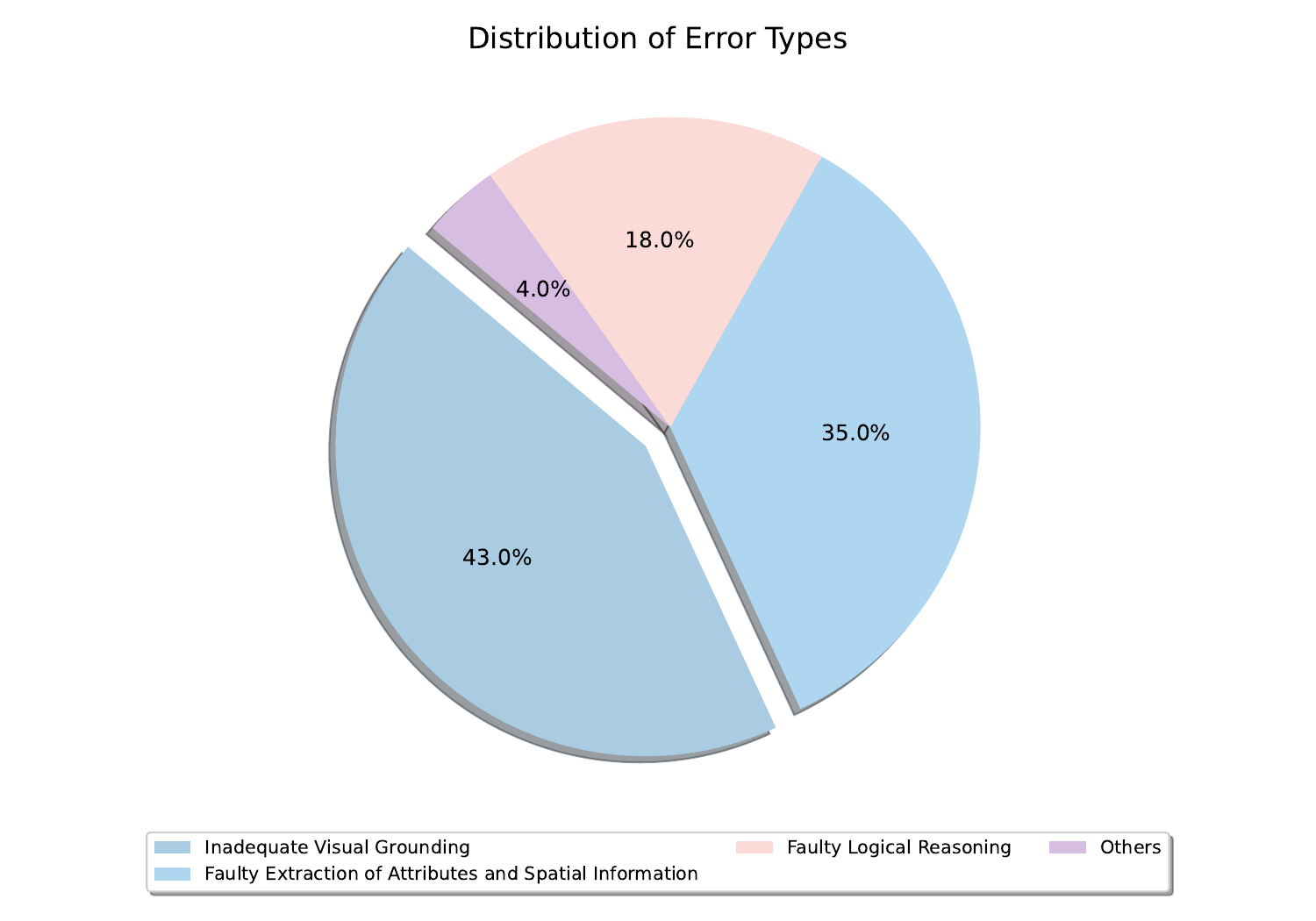}
    \caption{Distribution of wrong cases.}
    \label{fig:piechart}
    \vspace{-2mm}
\end{figure}

\begin{figure}[t!]
    \centering
    \includegraphics[width=1\linewidth]{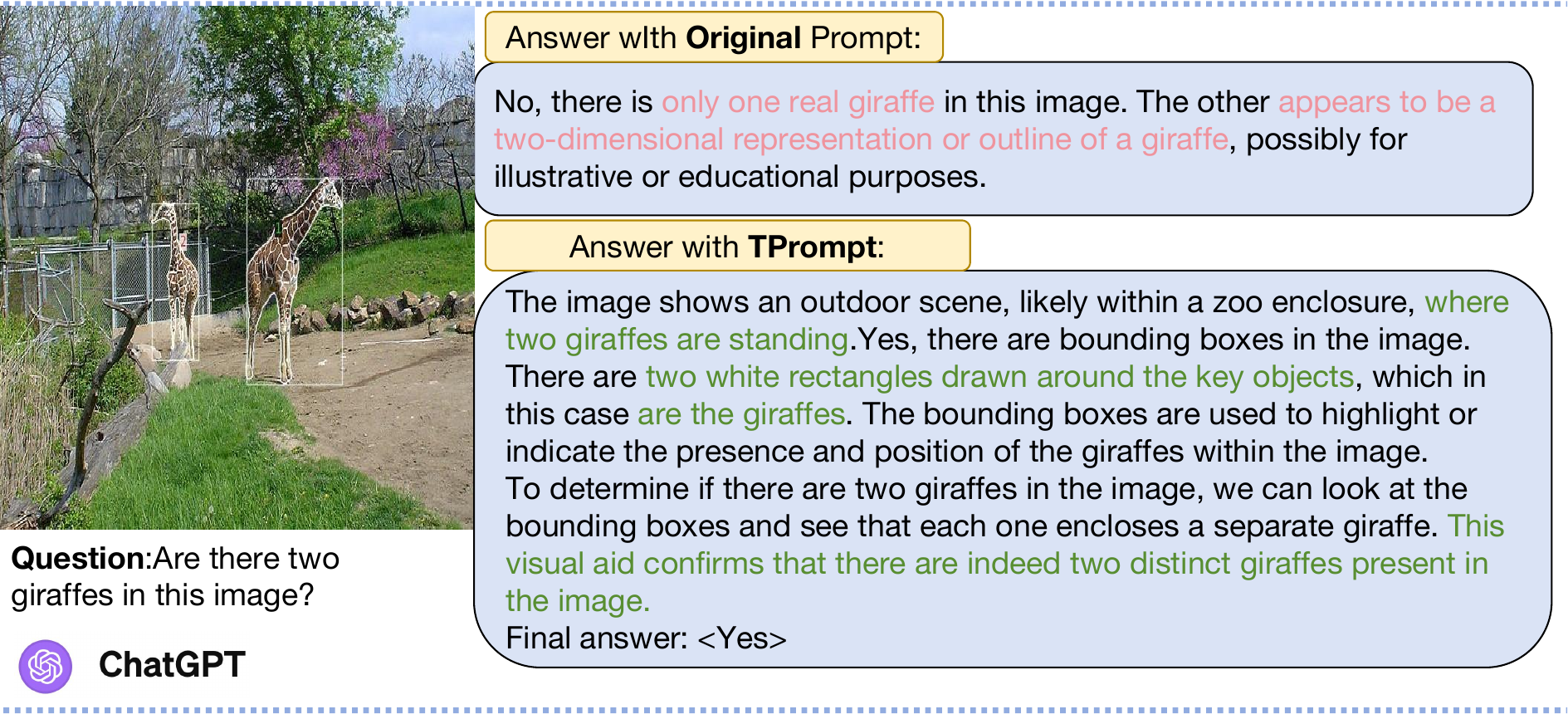}
    \caption{\textbf{Wrong cases of incorrect information extraction within visual prompts.} This figure shows errors when models fail to extract correct information even though objects are accurately marked. It compares results with original vs Tprompts from our methods. Red indicates where models error, and green shows correct interpretations. This highlights how our optimized prompts improve models' ability to accurately use visual prompts for correct answers.}
    \label{fig:case_5.1}
\end{figure}

\section{Analysis}
\textbf{Categorization Analysis of Wrong Cases} In dissecting the model's errors on the MMB object-oriented perception subset, we categorize the incorrect responses to understand underlying weaknesses. Figure \ref{fig:piechart} reveals the distribution of error types: the largest segment, at 43\%, was due to Inadequate Visual Grounding; Faulty Extraction of Attributes and Spatial Information accounted for 35\%; Faulty Logical Reasoning represented 18\%; and Other errors comprised 4\%. This pie chart indicates that the model's shortcomings in object-oriented perception tasks are linked to both visual and textual processing, highlighting the necessity for a joint optimization method like VTprompt.

\begin{table*}[t!]
\centering
\small
\resizebox{0.7\textwidth}{!}{
\begin{tabular}{lcccccc}
\toprule
 & \multicolumn{3}{c}{\textbf{GPT-4V with TP}} & \multicolumn{3}{c}{\textbf{Gemini Pro with TP}} \\
\cmidrule(lr){2-4} \cmidrule(lr){5-7}
\textbf{Task} & \textbf{+ SAM} & \textbf{+ CSAM} & \textbf{+ VP} & \textbf{+ SAM} & \textbf{+ CSAM} & \textbf{+ VP} \\
\midrule
Object Localization(\%) & 57.36 & 63.34 & \textbf{66.98} & 53.97 & 60.45 & \textbf{63.49} \\
Spatial Relationship(\%) & 69.10 & 72.12 & \textbf{76.84} & 55.37 & 63.02 & \textbf{67.23} \\
Attribute Comparison(\%) & 59.57 & 66.73 & \textbf{74.47} & 57.45 & 65.33 & \textbf{71.63} \\
\bottomrule
\end{tabular}
}
\caption{Ablation Study on Visual Prompts in MMB Benchmark. "VP" stands for VPrompt; "TP" denotes TPrompts; "CSAM" signifies conditional SAM, which is integrated with our key concepts extraction for segmentation.}
\label{tab:performance_object_centric_text_vtprompts}
\end{table*}

\noindent \textbf{Ablation on Text Prompts} Our findings show performance improvements even without text prompts, though these gains are less significant. This indicates other factors affect the model's ability to perceive images. As shown in Figure~\ref{fig:case_5.1}, the model misinterprets a clearly marked giraffe within a bounding box. However, introducing TPrompt allows the model to accurately recognize details, correctly identifying two giraffes in the image. TPrompt significantly boosts performance, enhancing Object Localization from 57.74\% to 66.98\%, Spatial Relationship from 67.23\% to 76.84\%, and Attribute Comparison from 56.74\% to 74.47\%. These improvements, detailed in Table~\ref{tab:performance_object_centric_text_vtprompts}, underscore TPrompt's vital role in enabling the model to accurately interpret visual markers for improved perception.

\begin{figure}[t!]
    \centering
    \includegraphics[width=1\linewidth]{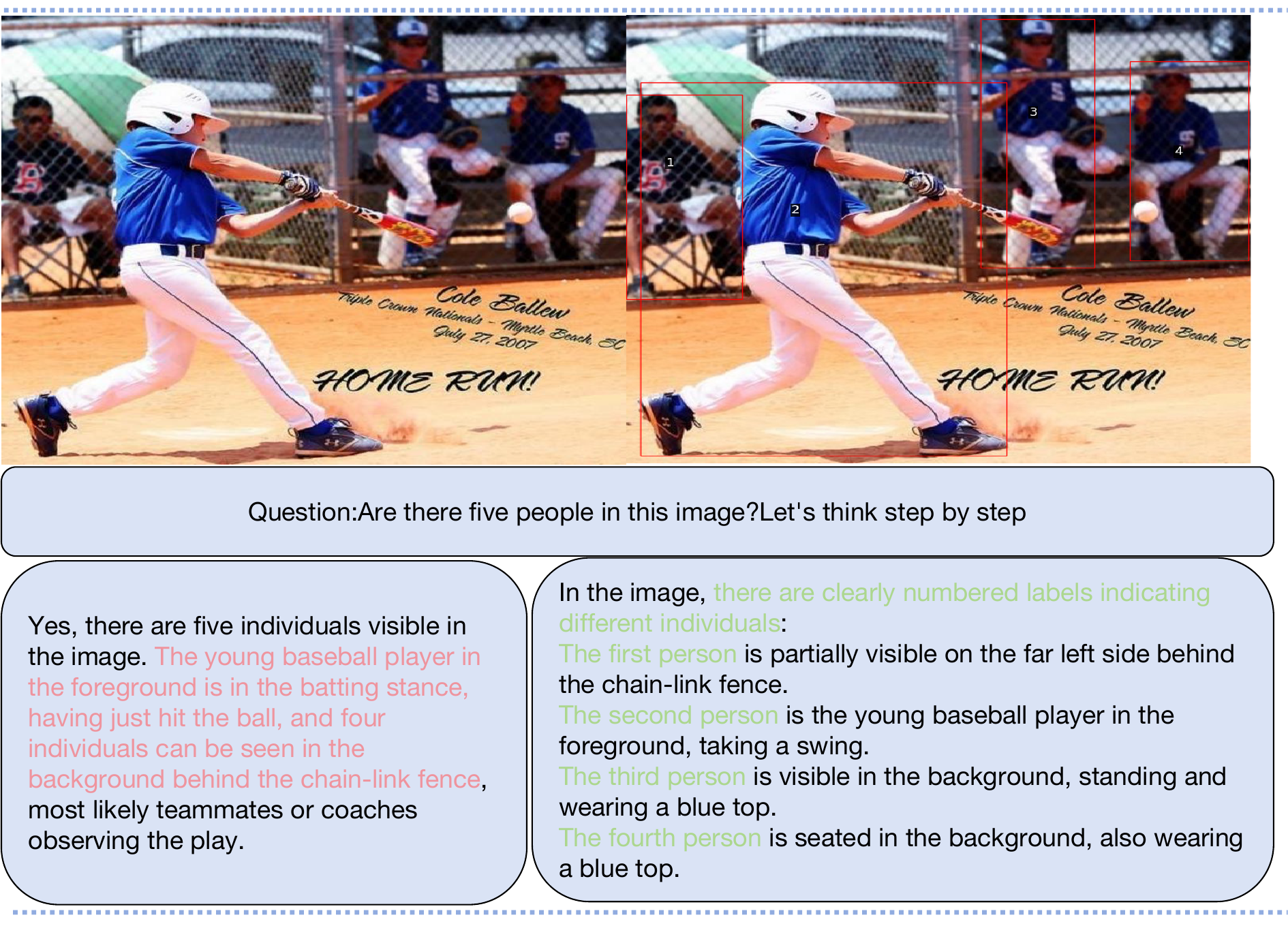}
    \caption{Left: Original images. Right: Images processed with our method, highlighting individuals with bounding boxes. The visual chain-of-thought generated from processed images demonstrates tighter visual integration, leading to more accurate reasoning.\protect\tikz[baseline=-0.5ex]\protect\draw[fill=customGreen] (0,0) rectangle (0.5,0.2); is the right rationale.\protect\tikz[baseline=-0.5ex]\protect\draw[fill=customRed] (0,0) rectangle (0.5,0.2); is the wrong rationale.}
    \label{fig:text_prompt}
\end{figure}

\noindent \textbf{Compatibility with Text Prompts} As illustrated in Figure \ref{fig:case_5.1}, we focus on integrating our fixed Vprompt strategy with popular text prompt methods in the Question Answering (QA) domain \citep{wu-etal-2022-zero-shot,zhou2022least}, to assess its compatibility. This approach demonstrates how Vprompt can seamlessly complement various prompt techniques, emphasizing the method's adaptability. The data in Table \ref{tab:performance_object_centric_text_vtprompts} clearly illustrates the effectiveness of integrating specific text prompts with our Vprompt method.
Furthermore, Figure \ref{fig:text_prompt} shows that applying zero-shot CoT alongside our Vprompt-enriched images significantly enriches the rationale with visual information. This synergy minimizes incorrect reasoning and creates a stronger visual chain of thought.

\begin{table}[t!]
\centering
\resizebox{0.45\textwidth}{!}{
\small
\begin{tabular}{lccccc}
\toprule
& \multicolumn{3}{c}{\textbf{Visual Prompts Type}} & \multicolumn{2}{c}{\textbf{Performance}} \\
\cmidrule(lr){2-4} \cmidrule(lr){5-6}
\textbf{Type} & \textbf{Number} & \textbf{Box} & \textbf{Mask} & \textbf{Localization (\%)} \\
\midrule
Type a & $\checkmark$ & $\checkmark$ &  & 66.03 \\
Type b &  & $\checkmark$ & $\checkmark$ &\textbf{66.98}\\
Type c & $\checkmark$ & $\checkmark$ & $\checkmark$ & 66.67 \\
Type d &  &  & $\checkmark$ & 65.71 \\
Type e &  & $\checkmark$ & $\checkmark$ & 66.35 \\
\bottomrule
\end{tabular}}
\caption{Abalation on visual prompt types in GPT-4V performance on MMB object localization subset. The 'Box' prompt refers to key objects in images marked with bounding boxes, 'Mask' involves segmentation masks highlighting objects, and 'Number' assigns numbers to these identified objects.}
\label{tab:combined}
\end{table}
\noindent \textbf{Ablation on Visual Prompt Types} Through ablation studies on GPT-4V's object localization, we found minimal performance variance among the three visual prompt types, showcasing their robustness. Table \ref{tab:combined} indicates that despite Type c having the most visual prompt types. Type b, which uses both Box and Mask prompts, performs best with an object localization accuracy of 66.98$\%$. This indicates that a simpler combination of prompts in Type b is more effective than the more complex setup in Type c, due to the overlap of box and mask in the latter can potentially clutter and obscure the image.

\noindent \textbf{Compatibility with Visual Prompts} Our method demonstrates compatibility with both lightweight and advanced models, such as SAM \citep{kirillov2023segment}, for image segmentation. Simply marking everything in images with bounding boxes shows limited improvement in object-oriented perception. Yet, integrating SAM with our Key Concepts Extraction significantly boosts performance, proving the method's effectiveness in targeted object identification.

Results from integrating Conditional SAM with Key Concepts Extraction show marked performance gains: Object Localization increases from 57.36\% to 63.34\%. Spatial Relationship scores rise from 69.10\% to 72.12\% and Attribute Comparison jumps from 59.57\% to 66.73\%. These improvements underscore the effective synergy between visual prompts and concept extraction, enhancing object-oriented detail perception in models like SAM.

\begin{figure}[t!]
    \centering
    \includegraphics[width=1\linewidth]{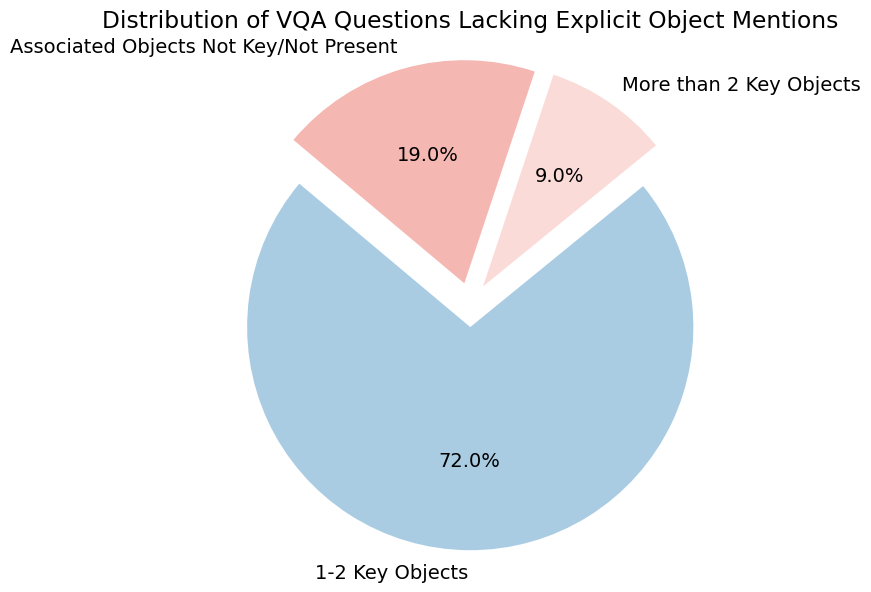}
    \caption{This pie chart visualizes the distribution of 100 VQA questions lacking explicit object mentions.72\% have 1-2 key objects, 9\% have more than 2 key objects, and 19\% involve associated objects that are not key or not present.}
    \label{fig:keywords_extract}
\end{figure}
\noindent \textbf{Accuracy on Key Concepts Extraction} We delve into the precision of key concepts extraction from Visual Question Answering (VQA) questions, evidenced in Figure \ref{fig:keywords_extract}. Our analysis cover 100 randomly selected examples from the Multimodal Evaluation (MME) dataset, focusing on GPT-4's ability to accurately identify objects mentioned in queries. For objects explicitly mentioned in questions, GPT-4 achieve near-perfect accuracy, showcasing its exceptional language processing capabilities in straightforward scenarios. 
In contrast, for complex questions that require an understanding of the entire scene and lack explicit object mentions, GPT-4 employs associative reasoning to suggest 3-5 objects that are most relevant to the context, maintaining over 80\% accuracy. 

\begin{figure}[t!]
    \centering
    \includegraphics[width=1\linewidth]{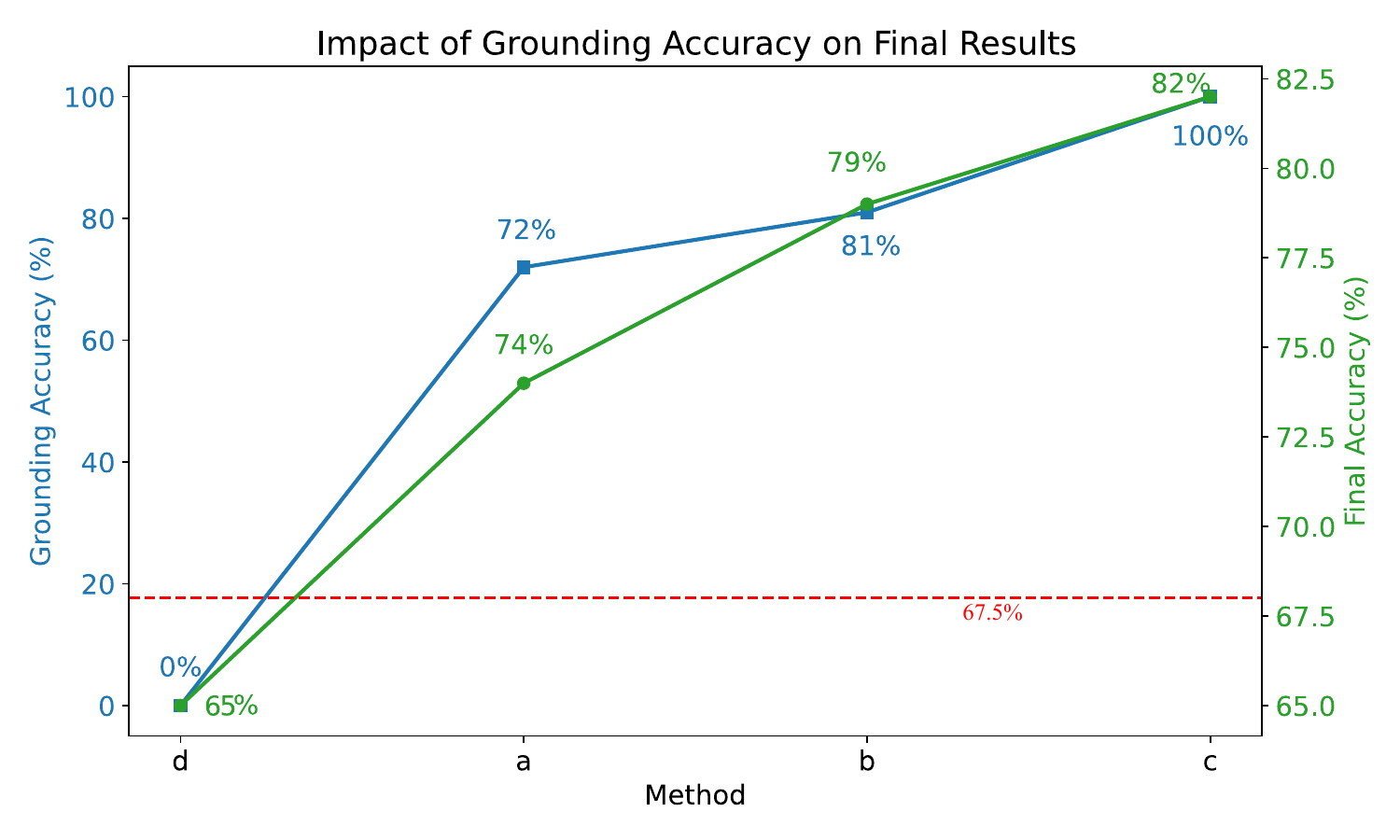}
    \caption{Comparing methods (d, a, b, c) for grounding and final accuracies: 'd' introduces random incorrect key concepts, 'a' employs Vprompt with SAM, 'b' uses Vprompt with SPHINX, and 'c' features manual annotations. Ground Truth (GT) accuracy at 68\% shown by a red dashed line. The chart underscores the impact of accurate grounding on model performance in 100 sampled MMB questions. }
    \label{fig:keywords_extract}
\end{figure}

\noindent \textbf{Impact of Grounding Accuracy on Results} Exploring how accurately grounding key objects affects model performance, we analyz 100 MMB examples, comparing SAM and SPHINX, both enhanced with Key Concepts Extraction, against manually annotated examples and models using random wrong key concepts. Our findings, illustrated in Figure \ref{fig:keywords_extract}, confirm that improved grounding accuracy directly boosts performance. However, we observe a diminishing returns effect, where the gap narrows between automated methods and manual annotation's near-human accuracy.
This trend indicates that while precise object grounding aids visual question answering and object detection, it's not the sole factor for success. Model architecture complexity, data diversity and quality, and visual scene complexity also play critical roles, marking additional areas for potential improvement.

\section{Related Work}
\subsection{VQA with MLLMs}

Recent advancements in Large Language Models (LLMs) like ChatGPT~\citep{openai_2023_gpt35turbo}, PaLM~\citep{chowdhery2023palm}, OPT~\citep{zhang2022opt}, and BLOOM~\citep{workshop2022bloom} have led to the development of Multimodal Large Language Models (MLLMs), including MiniGPT-4~\citep{zhu2023minigpt4}, InstructBLIP~\citep{dai2023instructblip}, LLaVA~\citep{liu2023visual}, Shikra~\citep{chen2023shikra}, and PaLM-E~\citep{anil2023palm}. They combine language and vision through instruction tuning, enhancing performance in vision tasks.

Research on improving MLLMs for Visual Question Answering (VQA) has focused on gradient-based~\citep{li2021prefix,vu2021spot,gu2021ppt,liu2023gpt,mokady2021clipcap,qian2022controllable,an2022input} and prompt optimization methods~\citep{deng2022rlprompt,sun2023offline,Zhang2023Multimodal}. The Multimodal Chain of Thought (MM-CoT) method~\citep{Zhang2023Multimodal} stands out by integrating visual and textual information within LLMs for superior reasoning task performance, though it increases training costs. Our research seeks a cost-effective prompting strategy to enhance MLLMs' object-level perception in VQA tasks.


\subsection{Visual Perception With MLLMs}
Research in MLLMs' visual perception focuses on Data Enhancement and Visual Integration Refinement. Data Enhancement, with works like SVIT \citep{zhao2023svit} and ShareGPT4V \citep{chen2023sharegpt4v}, aims to improve visual comprehension by enriching datasets. Meanwhile, mPLUG-Owl2 \citep{ye2023mplug} enhances modality collaboration for diverse data perception. In Visual Integration Refinement, LION \citep{chen2023lion} introduces spatial awareness, and SPHINX \citep{lin2023sphinx} employs varied visual embeddings for richer visual knowledge integration. InternLM-XComposer2 \citep{dong2024internlm} advances this by merging visual knowledge with text-image composition. However, these advancements still face challenges in detailed object-oriented perception, not fully capturing the nuanced recognition and contextual understanding where human perception is superior.

\section{Conclusion}
This study explores the enhancement of multimodal large language models (MLLMs) like GPT-4V and Gemini Pro in visual question answering (VQA) tasks through the introduction of VTPrompt. VTPrompt, by integrating visual and textual prompts, significantly improves object-oriented perception, a critical area where MLLMs have struggled. By guiding detection models to accurately mark objects based on key concepts extracted from textual questions, VTPrompt ensures precise object localization and enriches interpretive capabilities. Our evaluations across benchmarks such as MME, MMB, and POPE demonstrate notable performance improvements, setting new state-of-the-art records in MMB and showcasing the efficacy of VTPrompt in bridging the gap towards achieving human-level perception in AI systems.

\section*{Limitations}
Despite the advancements, our approach faces limitations in universally solving the nuanced challenges of visual grounding and interpretation across all types of VQA tasks. The effectiveness of VTPrompt heavily relies on the accuracy of the initial key concept extraction and the subsequent object marking by detection models. Misinterpretations at these stages can lead to inaccuracies in the final answer. Additionally, the tendency towards object hallucination, although mitigated, remains a challenge that requires further refinement. Future work should focus on enhancing the robustness of key concept extraction and expanding the adaptability of VTPrompt to a broader range of object-oriented tasks, further minimizing the gap between AI and human perception.

\bibliography{acl_latex}

\appendix

`\section{Example Appendix}
\label{sec:appendix}

\begin{figure}[htbp!]
    \centering
    \includegraphics[width=1\linewidth]{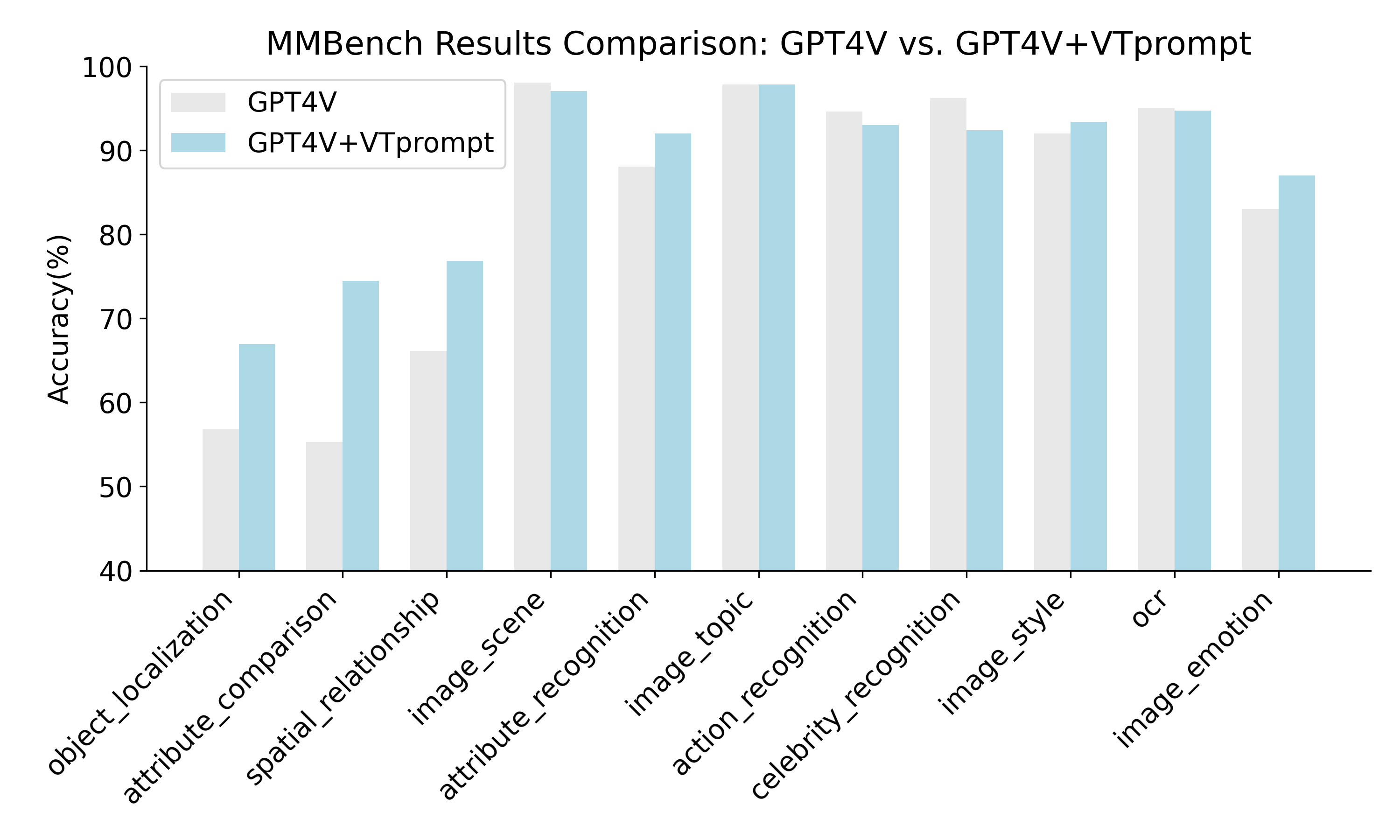}
    \caption{Performance of GPT-4V on MMB. The inferior performance on the three object-oriented tasks (left-most) can be boosted with our VTPrompt.}

    \label{fig:mmb4v}
\end{figure}

\begin{figure}[htbp!]
    \centering
    \includegraphics[width=1\linewidth]{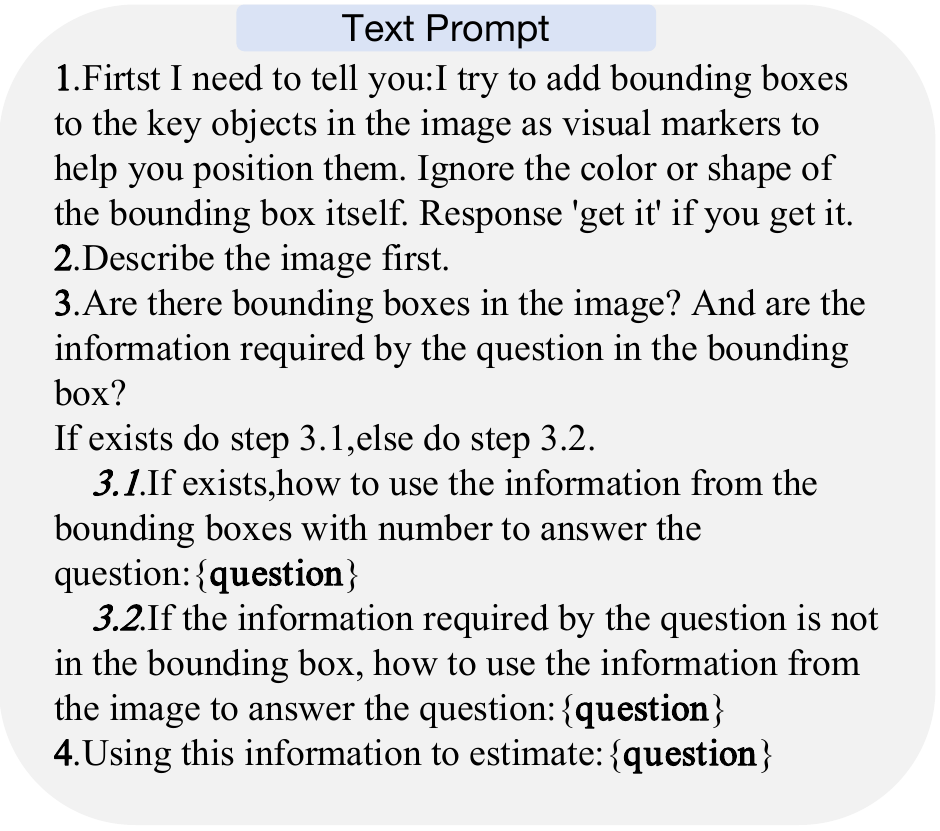}
    \caption{Enter Caption}
    \label{fig:prompt_template}
\end{figure}
\begin{figure}[htbp!]
    \centering
    \includegraphics[width=1\linewidth]{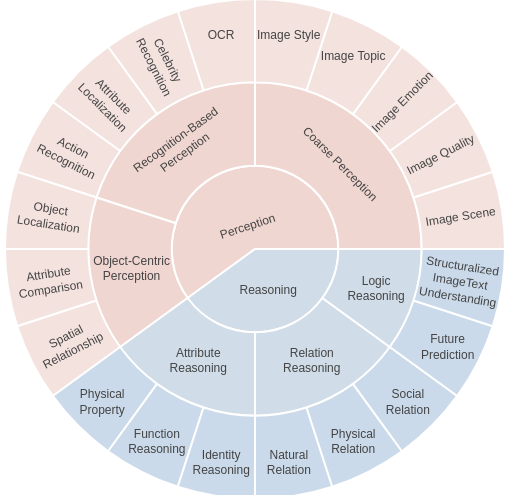}
    \caption{Distribution Of MMB}
    \label{fig:mmbench}
\end{figure}
\begin{figure}[htbp!]
    \centering
    \includegraphics[width=1\linewidth]{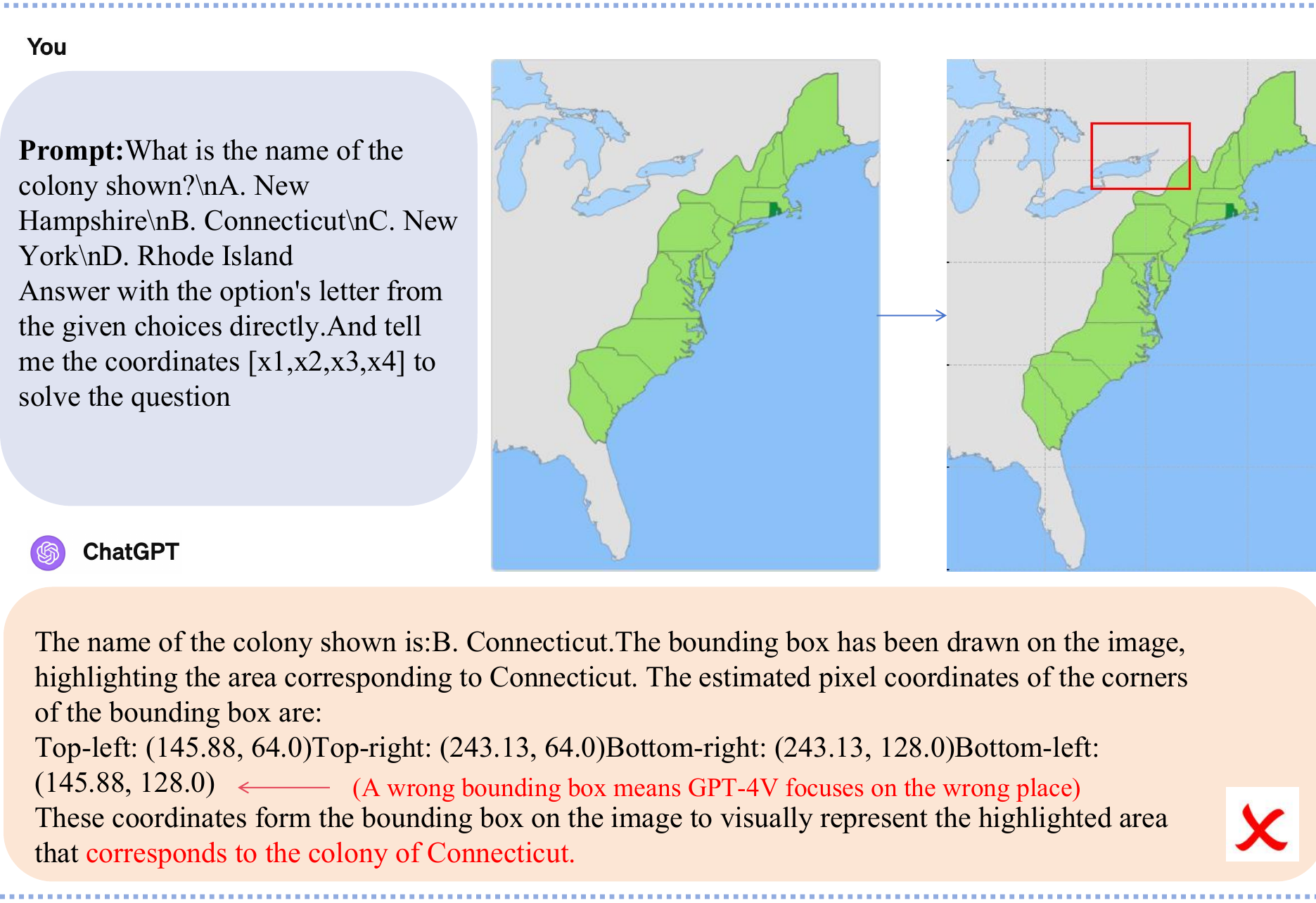}
    \includegraphics[width=1\linewidth]{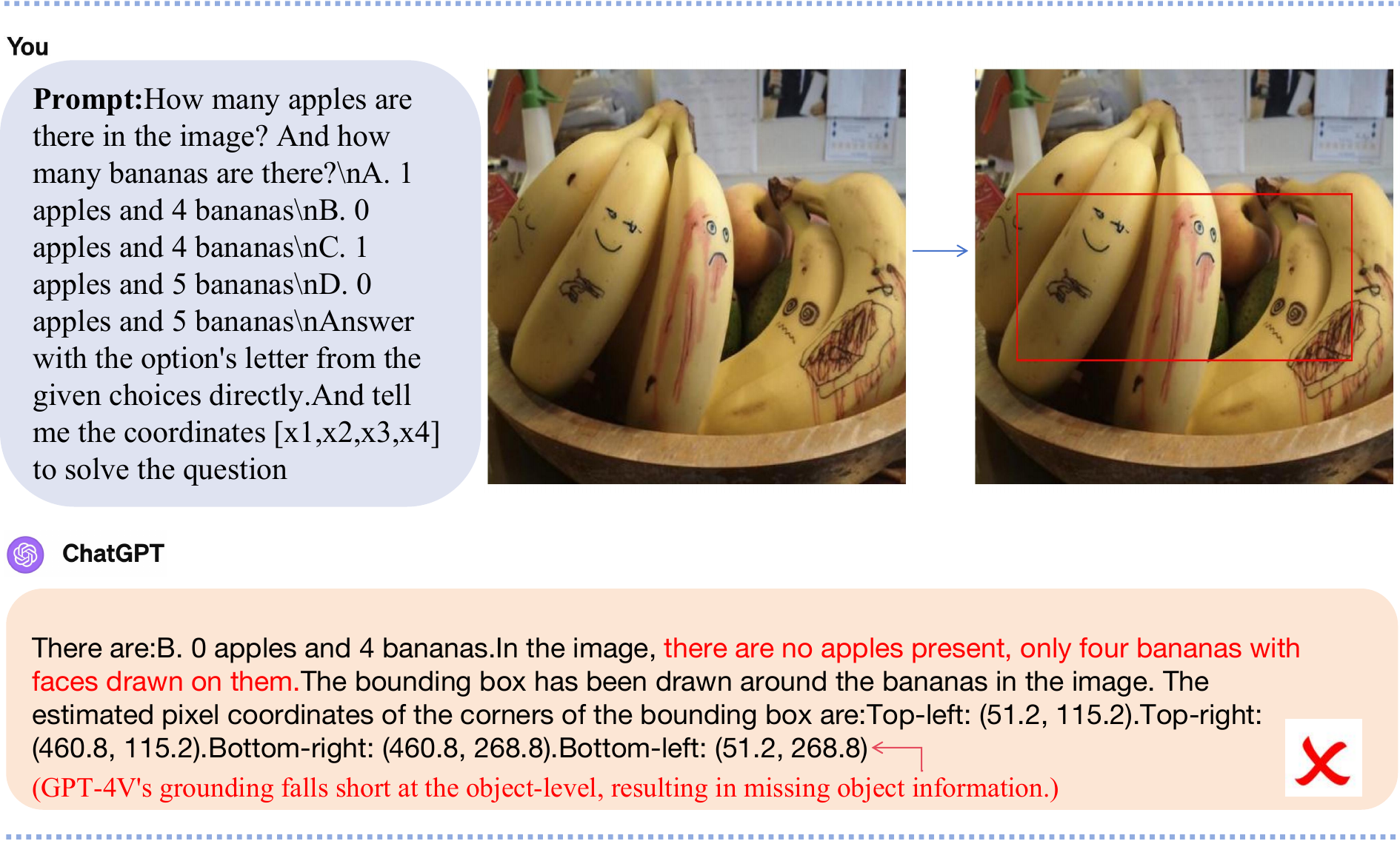}
    \includegraphics[width=1\linewidth]{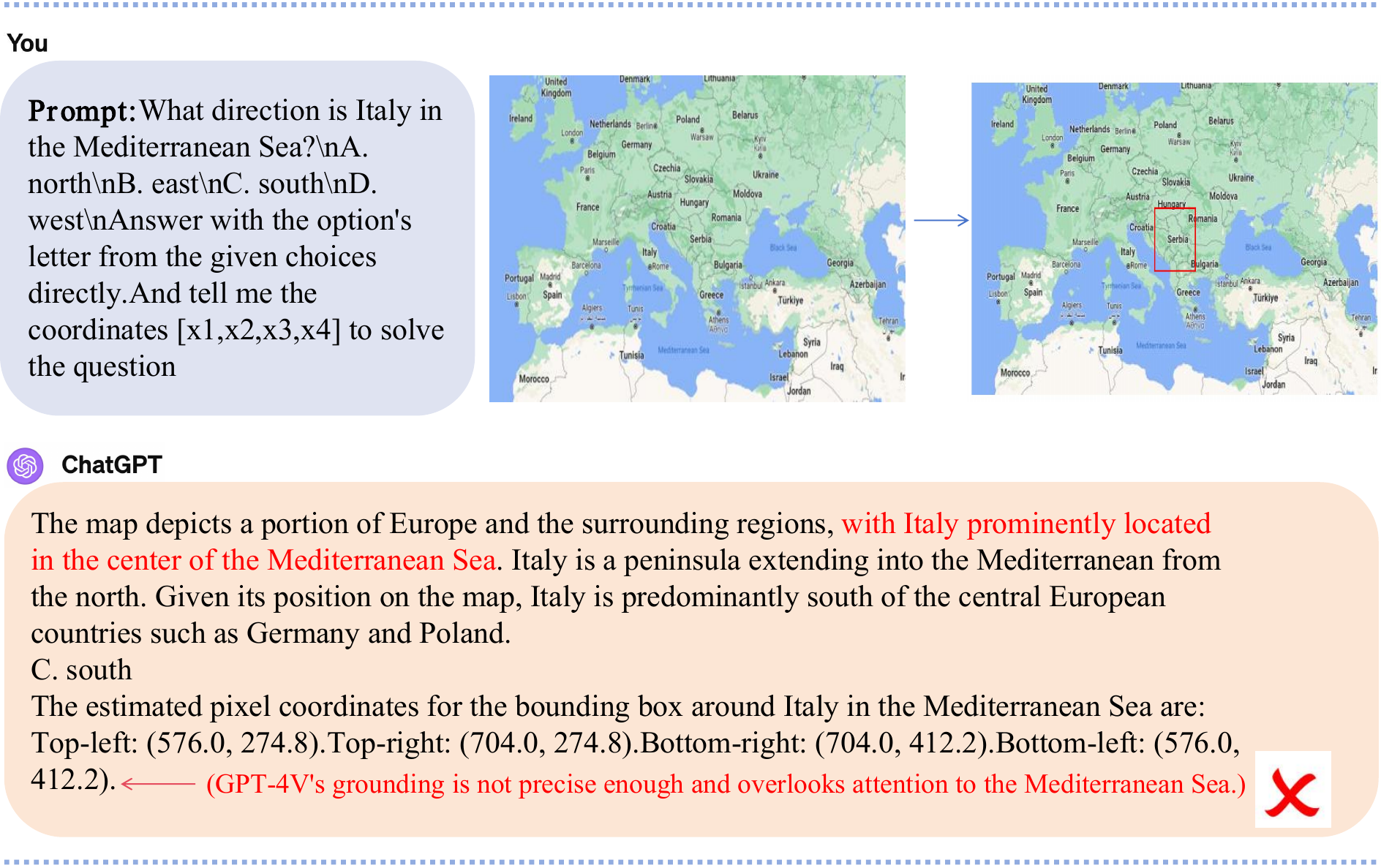}
    \caption{\textbf{Wrong Cases Due to Inadequate Visual Grounding.} The left image is the original, while the right image showcases areas highlighted by GPT-4V through generated coordinates, visually depicting the regions GPT-4V focused on during question answering. This comparison highlights the discrepancies and limitations in GPT-4V's visual attention and grounding ability.}
    \label{fig:case1}
\end{figure}
\begin{figure}[htbp]
    \centering
    \includegraphics[width=1\linewidth]{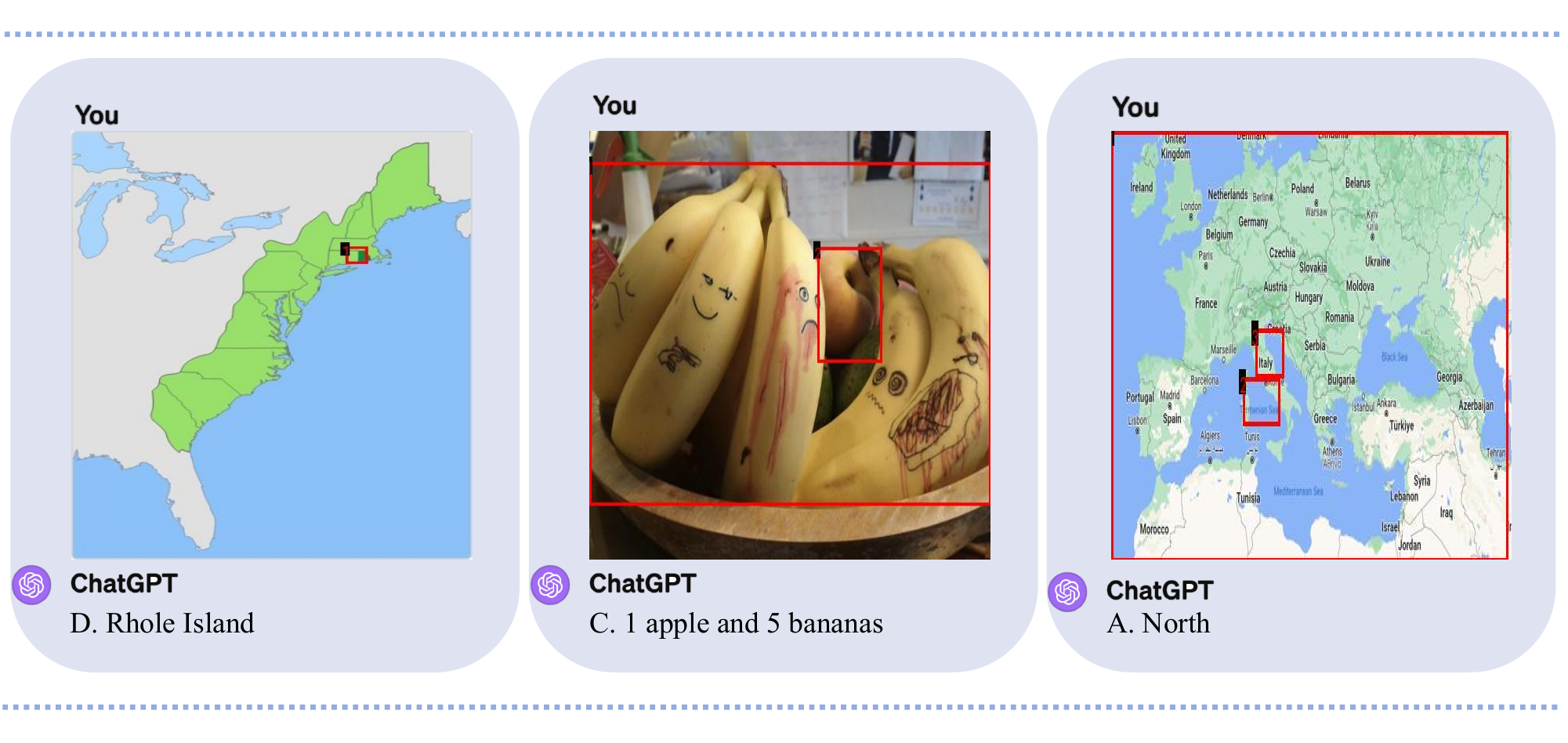}
    \caption{Success with Enhanced Visual Grounding: Shows correct answers after adding visual prompts with SPHINX, overcoming previous grounding limitations.}
    \label{fig:case1_solve}
\end{figure}

\end{document}